\documentclass[10pt,twocolumn,letterpaper]{article}

 \usepackage{cvpr}              % To produce the CAMERA-READY version
\usepackage{adjustbox}
\usepackage{multirow}
\usepackage{tabularx}
\usepackage{float}
\definecolor{cvprblue}{rgb}{0.21,0.49,0.74}
\usepackage[pagebackref,breaklinks,colorlinks,allcolors=cvprblue]{hyperref}

\def\paperID{*****} % *** Enter the Paper ID here
\def\confName{CVPR}
\def\confYear{2026}

\title{The 1st PortraitCraft Challenge: A CVPR 2026 Workshop Competition on Portrait Composition Understanding and Generation}

\author{
Zijie Lou\textsuperscript{\textdagger} \and
Youyun Tang\textsuperscript{\textdagger} \and
Xiaochao Qu\textsuperscript{\textdagger} \and
Haoxiang Li\textsuperscript{\textdagger} \and
Ting Liu\textsuperscript{\textdagger} \and
Luoqi Liu\textsuperscript{\textdagger} \and
Xun Zhu \and
Zheng Zhang \and
Xi Chen \and
Miao Li \and
Ji Wu \and
Dizhe Zhang \and
Xian Ge \and
Sujia Wang \and
Ruiyang Zhang \and
Jiaming Wang \and
Xianshun Wang \and
Lu Qi \and
Boao Kang \and
Wei Zhou \and
Jinghui Sun \and
Zhenyu Yan \and
Jiliang Zhao \and
Rui Yang \and
Yipo Huang \and
Boyuan Liu \and
Shanglin Li \and
Zifan Xie \and
Yichen Zhang \and
Anlan Wang \and
Wenfeng Lin \and
Mingyu Guo  \and
Dong Li  \and
Xinghao Wang  \and
Yanting Li  \and
Shanzhao Tong \and
Shuai He \and
Qiu Zhou \and
Yongqi Yang \and
Taoyang Mu \and
Dianqiao Lei \and
Anlong Ming \and
Huadong Ma 
}

\begin{document}
\maketitle

\footnotetext[1]{Authors marked with \textsuperscript{\textdagger} are challenge organizers; all other authors are participants.}

\begin{abstract}
This paper presents an overview of the inaugural PortraitCraft Challenge, held as one of the official competitions at CVPR 2026. The challenge focuses on portrait composition understanding and generation, aiming to advance AI research in portrait aesthetics analysis and controllable image synthesis. Unlike existing datasets and tasks that primarily focus on global aesthetic scoring, PortraitCraft introduces a unified evaluation framework comprising two complementary tracks. Track 1 requires models to perform structured portrait composition understanding, including overall composition score prediction, fine-grained attribute-level judgment over 13 composition attributes, and image-grounded visual question answering. Track 2 requires models to generate portrait images from structured composition descriptions under explicit compositional constraints, with evaluation centered on composition alignment rather than conventional image similarity metrics. To support the challenge, we constructed and publicly released a large-scale portrait composition dataset consisting of approximately 50,000 curated real portrait images, providing multi-level supervision including global composition scores, fine-grained attribute annotations, attribute-level explanation texts, VQA (Visual Question Answering) pairs, and structured composition descriptions for generation. The challenge ran for nearly two months and attracted a total of 295 teams from universities, research institutions, and industry worldwide. This report describes the challenge setup, evaluation protocols, dataset composition, and final results, along with an analysis of the technical characteristics of the submitted solutions. The PortraitCraft Challenge provides a standardized and reproducible platform for research on portrait composition understanding and generation, and is expected to foster further progress in the fields of portrait aesthetics and controllable image generation.
\end{abstract}    
\section{Introduction}
Portrait composition is a key determinant of aesthetic quality in portrait photography and AI-generated content, yet it remains underexplored in structured vision benchmarks. Unlike generic image aesthetics, portrait composition involves subject-centered factors such as body pose, facial expression, spatial organization, and background interaction~\cite{cao2026artimuse,yang2022personalized,zhao2025can}. Existing datasets (e.g., AVA~\cite{gu2018ava}, AADB~\cite{kong2016photo}, APDDv2~\cite{jin2024apddv2}) primarily focus on coarse aesthetic scoring or artistic images, lacking structured supervision for portrait-centered composition analysis and generation.

To bridge this gap, we recently introduced PortraitCraft~\cite{sha2026portraitcraft}, a large-scale benchmark comprising ~50,000 curated real portrait images with multi-level annotations, including global composition scores, 13 fine-grained attributes, VQA pairs, and structured composition descriptions for generation.

Building upon this benchmark, we organized the 1st PortraitCraft Challenge as an official CVPR 2026 workshop competition. The challenge features two complementary tracks:
\begin{itemize}
\item{Track 1 (composition understanding): overall score prediction, attribute-level judgment, and VQA.}

\item{Track 2 (composition generation): generating portraits from structured composition descriptions, evaluated by composition alignment.}
\end{itemize}
The challenge attracted 295 teams worldwide and ran from April to May 2026. This report summarizes the challenge setup, evaluation protocols, and final results, followed by an analysis of top-performing solutions.
\section{Challenge Format, Evaluation Metrics, and Ranking}
The 1st PortraitCraft Challenge was organized as an official CVPR 2026 workshop competition based on the PortraitCraft benchmark~\cite{sha2026portraitcraft}. The challenge comprised two independent tracks: Track 1 – Portrait Composition Understanding and Track 2 – Portrait Composition Generation. Both tracks were hosted on CodaBench.\footnote{Track 1: \url{https://www.codabench.org/competitions/15386/}; Track 2: \url{https://www.codabench.org/competitions/15479/}}

\subsection{Data Splits and Submission Phases}
Track 1 (Understanding) utilized the PortraitCraft dataset with 47,863 training and 2,000 test images. Training annotations included global composition scores, 13 attribute labels, attribute-level explanations, and VQA pairs. Participants submitted predictions on the test set via the CodaBench server. Track 2 (Generation) used a subset of 5,000 image–description pairs (4,500 training, 500 test). Participants generated portrait images from the test descriptions alone. The challenge ran for approximately six weeks (April 5 – May 17, 2026). After the final phase, each team was required to submit code, checkpoints, and a method description for validation.

\subsection{Evaluation Metrics and Ranking}
\subsubsection{Track 1: Portrait Composition Understanding}

Following the original benchmark~\cite{sha2026portraitcraft}, Track 1 evaluated models on three components:
\begin{itemize}
\item{Overall score prediction: SRCC and PLCC between predicted and expert-annotated scores.}

\item{Attribute-level judgment: Level accuracy across 13 composition attributes.}

\item{Visual question answering: QA accuracy on multiple-choice questions.}
\end{itemize}
A composite score was computed as the average of the normalized SRCC, PLCC, level accuracy, and QA accuracy. Teams were ranked by this composite score. The quantitative evaluation is provided in Tab.\ref{tab1}.

\subsubsection{Track 2: Portrait Composition Generation}
The evaluation was performed in two stages.

Stage 1 – Automatic scoring with Gemini-3.1-pro:
For each generated image, the Gemini-3.1-pro model produced two sub-scores (each 0–100):
\begin{itemize}
\item{Content score (weight 0.3): assesses visual plausibility, subject integrity, and absence of artifacts.}

\item{Composition score (weight 0.7): evaluates adherence to the input composition description (subject placement, spatial organization, visual center, etc.).}
\end{itemize}
The automatic total score was calculated as:
\begin{equation}
\text{AutoScore} = 0.3 \times \text{ContentScore} + 0.7 \times \text{CompositionScore}
\end{equation}

The initial leaderboard was ranked by AutoScore.

Stage 2 – Human expert validation for top-5 submissions:

For the five highest-ranked submissions on the leaderboard, every generated image was independently scored by three professional experience designers (0–100 scale, focusing on composition alignment and overall quality). The human score for a submission was the average of all image-level scores across the three designers.

The final ranking score was computed as:
\begin{equation}
\text{FinalScore} = 0.4 \times \text{AutoScore} + 0.6 \times \text{HumanScore}
\end{equation}
The final ranking of the challenge was determined by this FinalScore. The quantitative evaluation is provided in Tab.\ref{tab2}.

\begin{table*}[!t]
\centering
\caption{Quantitative evaluation of the solutions compared in the Final Phase for Track 1 (Portrait Composition Understanding). The top-3 teams submitted their official team names, and we have removed duplicate entries from the same team using different usernames. However, for ranks below the top-3, multiple entries from the same team under different usernames may still exist.}
\tabcolsep=15 pt
\label{tab1}
\begin{adjustbox}{width=\textwidth}
\begin{tabular}{c|cc|cccc|c}
\toprule[1pt]
Rank & Team & Username & SRCC ↑ & PLCC ↑ & Criteria Acc ↑ & Answer Acc ↑ & Total Score ↑ \\
\midrule
1 & MSIIP & ms & 0.923211 & 0.921970 & 0.764808 & 0.984500 & 0.898622 \\
2 & sky & sky & 0.930660 & 0.931163 & 0.732885 & 0.979500 & 0.893552 \\
3 & ZTE\_CHU & wilburzzz & 0.923499 & 0.921940 & 0.752538 & 0.969000 & 0.891744 \\
4 & - & abc3141106022 & 0.923439 & 0.921677 & 0.746846 & 0.943500 & 0.883866 \\
5 & - & uesrabcd & 0.922069 & 0.920580 & 0.752462 & 0.940000 & 0.883778 \\
6 & - & asuka123 & 0.922144 & 0.920622 & 0.752154 & 0.940000 & 0.883730 \\
7 & - & yanzy & 0.922028 & 0.920559 & 0.752192 & 0.940000 & 0.883695 \\
8 & - & qnmwan & 0.922012 & 0.920548 & 0.752192 & 0.940000 & 0.883688 \\
9 & - & lora11112222 & 0.922163 & 0.920633 & 0.751923 & 0.940000 & 0.883680 \\
10 & -  & wmyyl & 0.921945 & 0.920359 & 0.751577 & 0.940000 & 0.883470 \\
11 & - & monkey654 & 0.921878 & 0.920417 & 0.751269 & 0.940000 & 0.883391 \\
12 & - & lxy224 & 0.922681 & 0.921247 & 0.743462 & 0.942000 & 0.882348 \\
13 & - & aimer & 0.923150 & 0.921274 & 0.741923 & 0.940000 & 0.881587 \\
14 & - & abandonnn & 0.920206 & 0.918959 & 0.743462 & 0.939500 & 0.880532 \\
15 & - & jiangchao & 0.914296 & 0.914249 & 0.715038 & 0.976500 & 0.880021 \\
16 & - & vegetableman & 0.918267 & 0.916710 & 0.741538 & 0.939500 & 0.879004 \\
17 & - & wuqinzhuo & 0.913701 & 0.913685 & 0.712885 & 0.975500 & 0.878943 \\
18 & kk & zmhhhh & 0.917478 & 0.917325 & 0.714038 & 0.966000 & 0.878710 \\
19 & - & elgent123 & 0.922123 & 0.920612 & 0.732038 & 0.940000 & 0.878693 \\
20 & - & raodisa & 0.911468 & 0.911762 & 0.711885 & 0.975000 & 0.877529 \\
\bottomrule[1pt]
\end{tabular}
\end{adjustbox}
\end{table*}

\begin{table*}[!t]
\centering
\caption{Quantitative evaluation of the top-20 solutions for Track 2 (Portrait Composition Generation) in the Final Phase. The top-4 teams provided official names; duplicate entries from the same team with different usernames have been removed. For other ranks, such duplicates may still be present.}
\tabcolsep=8 pt
\label{tab2}
\begin{adjustbox}{width=\textwidth}
\begin{tabular}{c|cc|cc|c|c c}
\toprule[1pt]
Rank & Team & Username & Content Score ↑ & Composition Score ↑ & Auto Score ↑ & Human Score ↑ & Final Score ↑\\
\midrule
1 & ContentU\&G & dwayne & 87.8918 & 87.1383 & 87.3643 & 78 & 81.74572 \\
2 & elephant & elephant & 85.9820 & 84.0000 & 84.5946 & 68 & 74.63784 \\
3 & sky & sky & 87.1260 & 82.9740 & 84.2196 & 60 & 69.68784 \\
4 & PCU-vRobotit@BUPT & leonheb732 & 86.6940 & 83.5620 & 84.5016 & 58 & 68.60064 \\
5 & - & lklklk & 85.4267 & 82.3233 & 83.2543 & - & - \\
6 & - & wxhsdhr & 85.2345 & 82.2485 & 83.1443 & - & - \\
7 & - & leoleohaa & 85.4987 & 81.6519 & 82.8060 & - & - \\
8 & - & wxhwxh & 84.6220 & 81.7680 & 82.6242 & - & - \\
9 & - & asvka & 84.2980 & 81.7540 & 82.5172 & - & - \\
10 & - & donkey & 84.7893 & 81.2335 & 82.3003 & - & - \\
11 & Cyrus & cyrus\_smz & 84.9631 & 81.1365 & 82.2845 & - & - \\
12 & - & zeshanzhang & 84.8184 & 80.3760 & 81.7087 & - & - \\
13 & - & better & 84.9182 & 79.9528 & 81.4425 & - & - \\
14 & - & snowsnow & 83.8754 & 79.6677 & 80.9300 & - & - \\
15 & - & yuchen001 & 83.4878 & 79.3510 & 80.5920 & - & - \\
16 & - & qiu416 & 82.8960 & 78.4300 & 79.7698 & - & - \\
17 & - & wu\_tian\_ci & 84.6540 & 77.5940 & 79.7120 & - & - \\
18 & - & pp405 & 82.7980 & 78.3820 & 79.7068 & - & - \\
19 & - & t020202 & 83.2006 & 78.1497 & 79.6650 & - & - \\
20 & - & zyw729530 & 82.7940 & 78.2760 & 79.6314 & - & - \\
\bottomrule[1pt]
\end{tabular}
\end{adjustbox}
\end{table*}
\section{Methods}

\subsection{Track 1: Portrait Composition Understanding}
\subsubsection{Team MSIIP}

Team MSIIP trained Qwen3-VL models for portrait composition understanding and reported several preliminary observations that shaped their solution. First, they observed that changing the order of image/text or score/reason during training could affect the output, which they considered possibly related to the causal attention mechanism. Second, they found that models trained with discrete class labels and fine-grained regression scores showed different behaviors. The regression supervision appeared to provide finer gradient signals and relatively better ranking information, while the classification variant was more sensitive to exact decision boundaries.

\begin{figure}[t]
\centering
\includegraphics[width=\linewidth]{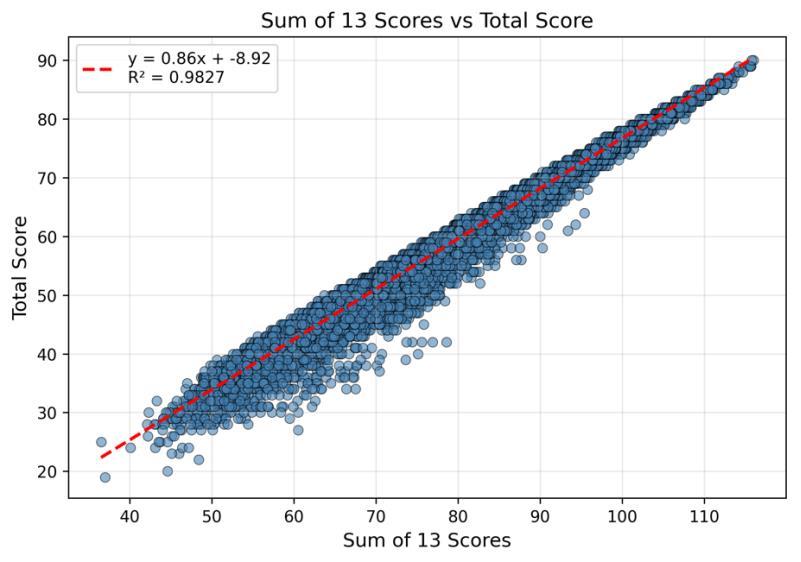}
\caption{The linear relationship between the sum of 13 scores and the total score reported by Team MSIIP.}
\label{fig:msiip_score_relation}
\end{figure}

The team also observed a strong linear association between the overall score and the sum of the 13 sub-scores on the training set, with $y = 0.86x - 8.92$ and $R^2 = 0.9827$, where $y$ denotes the overall score and $x$ denotes the sum of sub-scores. This relationship is shown in Fig.~\ref{fig:msiip_score_relation}. In addition, they reported that SRCC and PLCC for the overall score could exceed 0.9, while the per-criterion classification accuracy was noticeably lower. Based on this observation, they hypothesized that the model might rank sub-criteria reasonably well but might not follow the predefined thresholds precisely.

Considering model size limits, MSIIP trained two separate Qwen3-VL-4B variants. Model A was trained with the overall score, discrete class labels (A/B/C) for the 13 criteria, and the total score. Model B was trained with the overall score, fine-grained regression scores from 0.0 to 10.0, and the total score. This dual setup was used to capture both coarse categorical supervision and fine numerical signals.

For test-time augmentation, the team used horizontal flipping with corresponding directional word substitutions such as left/right and east/west, multiple resolutions including 448, 2048, and 8192, and diverse decoding by enabling \texttt{do\_sample} with tuned temperature. They also performed threshold calibration on the validation set. For each criterion, they searched for class boundaries that maximized classification accuracy against the ground-truth labels, instead of relying on default thresholds such as 5 and 7.

During inference, MSIIP combined predictions from the two models and all augmentations. The overall score was averaged across all forward passes, with an optional minor adjustment using the linear relationship between the overall score and the predicted sub-score sum. Class labels and VQA choices were obtained by majority voting across augmented outputs. For regression-based labels, sub-score predictions were first averaged over augmentations, and then the pre-calibrated thresholds were applied to produce the final Poor/Medium/Good categories.

\subsubsection{Team sky}

Team sky addressed the PortraitCraft Track-1 task of multi-dimensional portrait composition assessment by fine-tuning a 4B vision-language model, Qwen3.5-4B-VL, with LoRA. Their solution combines parameter-efficient supervised fine-tuning with knowledge distillation from a multi-model teacher ensemble and a 2-pass test-time augmentation ensemble at inference. The model jointly predicts 13 criteria scores, an overall aesthetic score from 1 to 100, and a 4-way VQA answer in a single forward pass.

The pipeline consists of three stages. First, knowledge distillation is performed from a multi-model teacher ensemble into a 4B student model via LoRA supervised fine-tuning. Second, multi-resolution and horizontal-flip test-time augmentation is used during inference, with predictions averaged in the continuous score space. Third, continuous criterion scores are mapped to discrete A/B/C levels using fixed thresholds after averaging.

For the model architecture, the team started from the multimodal VL variant of Qwen3.5-4B. The model was fine-tuned with LoRA targeting all linear projection layers. After training, the LoRA adapter was merged back into the base weights for efficient single-engine vLLM serving. The training labels were prepared using knowledge distillation from a multi-model teacher ensemble. Several larger Qwen variants were first fine-tuned independently on the official training set, and their predictions, including per-criterion continuous scores, total score, and VQA answer, were averaged in the continuous score space to produce ensemble soft labels. These ensemble labels were used as the training target for the 4B student instead of the raw original annotations.

\begin{figure}[t]
\centering
\includegraphics[width=\linewidth]{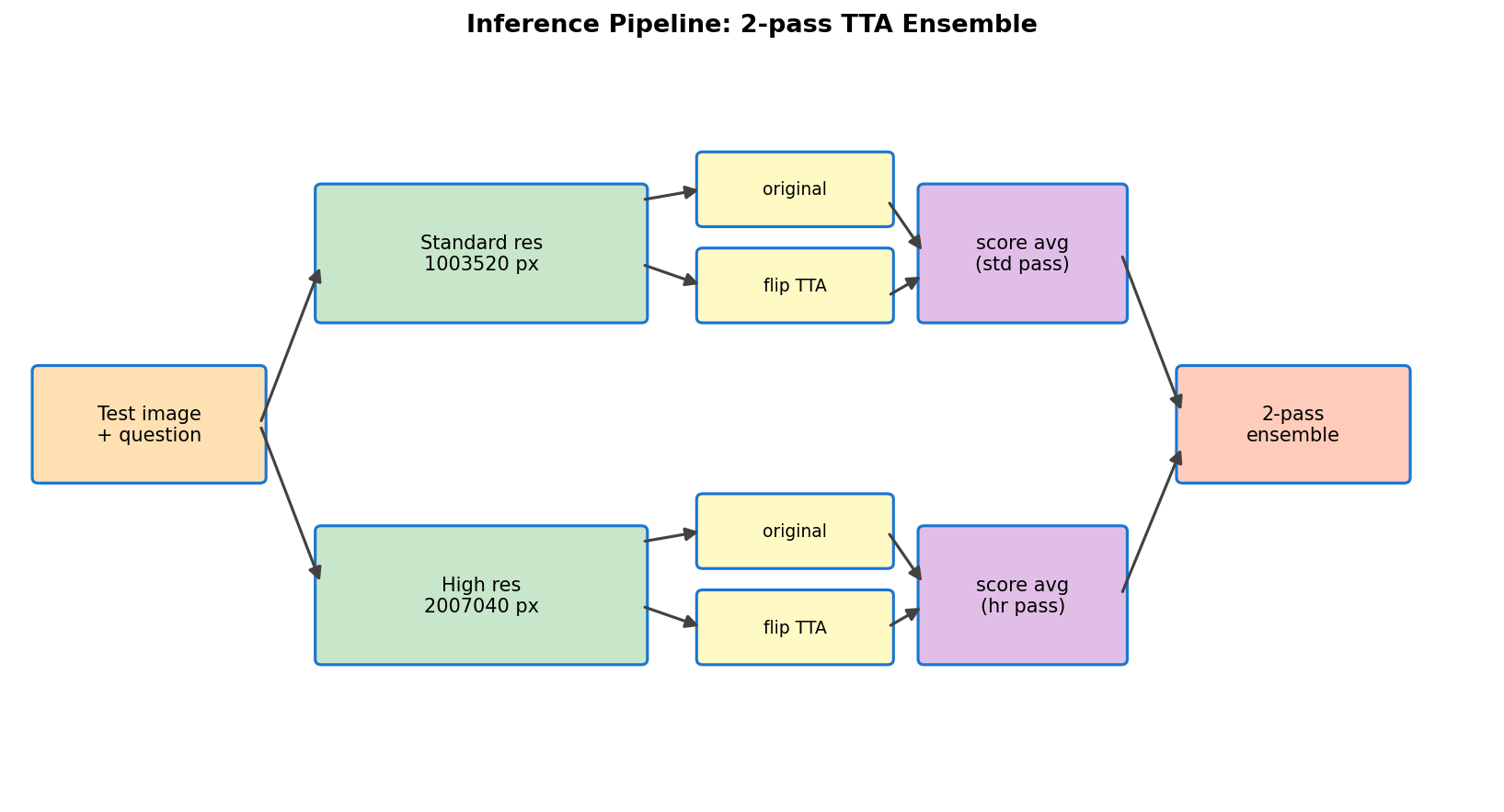}
\caption{Inference pipeline of Team sky. Standard- and high-resolution passes each apply horizontal-flip TTA, and per-pass averages are then averaged across passes.}
\label{fig:sky_inference_pipeline}
\end{figure}

During inference, the team used a 2-pass test-time augmentation ensemble, as shown in Fig.~\ref{fig:sky_inference_pipeline}. For each test image, the model was run four times in total, using two different input resolutions and both the original and horizontally flipped image. The standard-resolution pass used \texttt{max\_pixels = 1003520}, matching the training-time resolution, while the high-resolution pass used \texttt{max\_pixels = 2007040}, which is twice the training resolution. Within each resolution pass, predictions from the original image and the horizontally flipped image were averaged.

The final aggregation was performed in the continuous score space. For each of the 13 criteria, the final score was computed as the arithmetic mean of the four inference outputs. The overall \texttt{total\_score} was aggregated in the same way and rounded to an integer. The VQA answer was taken from the standard-resolution original pass, because some VQA questions encode spatial directionality. The submission package included merged model weights, an inference script, pinned dependency versions, and the team reported that training and inference were conducted on H20 GPUs.

\subsubsection{Team ZTE\_CHU}

Team ZTE\_CHU built their solution for Track 1 upon the Qwen3.5-VL-4B vision-language model, following the PortraitCraft task setting~\cite{sha2026portraitcraft}. The entire system consistently shares the same 4B-level base model, on top of which multiple lightweight LoRA adapters are trained to adapt to different prediction targets. During final inference, the team did not switch to a larger backbone model. Instead, they improved the overall performance through the complementary modeling capabilities of multiple LoRA adapters.

\begin{figure}[!t]
\centering
\includegraphics[width=\linewidth]{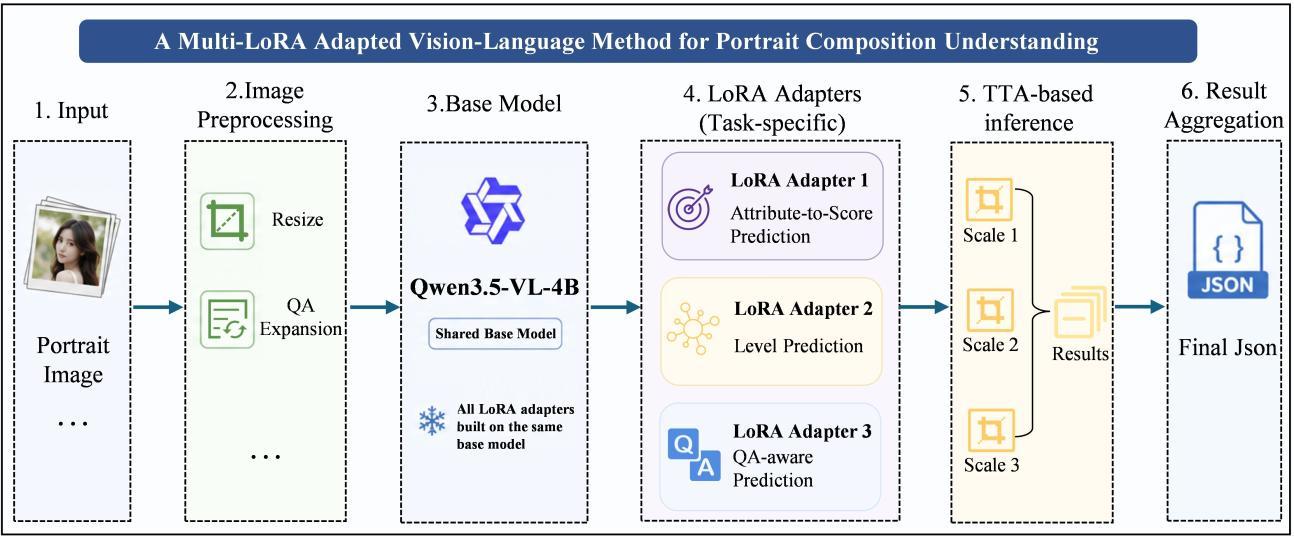}
\caption{Overall method workflow of Team ZTE\_CHU.}
\label{fig:zte_chu_workflow}
\end{figure}

To better model the different output formats required by Track 1, the team trained three LoRA adapters on the same Qwen3.5-VL-4B base model. The first LoRA mainly focuses on fine-grained composition attribute prediction and further uses the predicted attributes to assist the estimation of the overall composition score. The second LoRA adopts a Q-Align-style formulation and focuses on overall composition level prediction. The third LoRA is trained with samples containing QA questions, aiming to enhance the model's joint understanding of image content and question semantics. The overall workflow is shown in Fig.~\ref{fig:zte_chu_workflow}, and the LoRA adapter design is summarized in Tab.~\ref{tab:zte_chu_lora}.

\begin{table}[!t]
\centering
\caption{LoRA adapter design of Team ZTE\_CHU.}
\tabcolsep=15 pt
\label{tab:zte_chu_lora}
\begin{adjustbox}{width=\linewidth}
\begin{tabular}{lcc}
\toprule
Model & Parameters & Size \\
\midrule
Base(Qwen3.5-VL-4B) & 4,659,865,088 & 4.66B \\
LoRA Adapter 1 & 6,291,456 & 6.29M \\
LoRA Adapter 2 & 6,291,456 & 6.29M \\
LoRA Adapter 3 & 6,291,456 & 6.29M \\
Total(Base+3LoRA) & 4,678,739,456 & 4.68B \\
\bottomrule
\end{tabular}
\end{adjustbox}
\end{table}

\begin{table*}[!t]
\centering
\caption{Method variants and experimental observations reported by Team ZTE\_CHU.}
\label{tab:zte_chu_variants}
\begin{tabularx}{\textwidth}{l p{0.32\textwidth} X}
\toprule
Method Variant & Description & Observation \\
\midrule
SFT model &
Supervised fine-tuning using the constructed data. &
The structured output was significantly improved, and the model became better at learning composition attributes and a unified answering format. \\

SFT+Q-Align &
Discretize the overall composition score into ordered quality tokens, and obtain the final score through probability-weighted aggregation over the token probabilities. &
Score prediction became more stable, reducing the drift caused by direct numerical generation and providing complementary information to other LoRA adapters. \\

SFT + GRPO &
Add reward-based optimization on top of SFT. &
The consistency among the overall score, level judgments, and QA results was further improved. \\

Multi-LoRA system &
Complementary prediction by combining multiple LoRA adapters. &
Different prediction formulations complement each other, reducing the bias caused by a single modeling strategy. \\

TTA inference &
Use multi-scale input testing with resize ratios set to 0.95 / 1.0 / 1.05, combined with multi-configuration result aggregation. &
The multi-scale perspective reduces fluctuations caused by a single input scale and further improves the stability of the overall score and QA results. \\
\bottomrule
\end{tabularx}
\end{table*}

\begin{figure*}[!t]
\centering
\includegraphics[width=\textwidth]{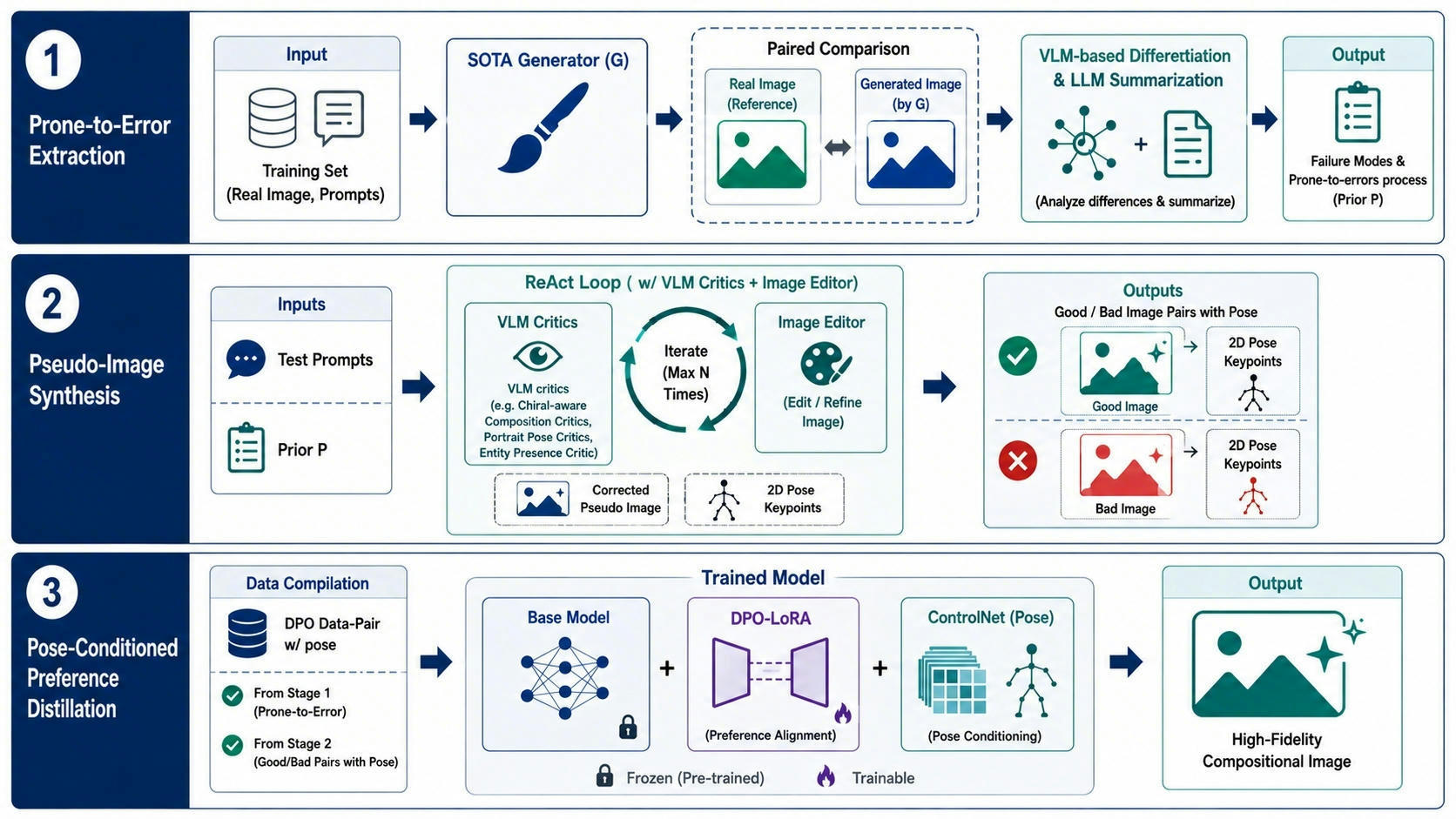}
\caption{Overview of the proposed three-stage pipeline for pose-conditioned preference distillation.}
\label{fig:contentug_pipeline}
\end{figure*}

The training process consists of two main stages. First, supervised fine-tuning is performed based on the constructed data, including QA questions, so that the model can learn the input-output format required by the track as well as the basic prediction ability. Then, GRPO reinforcement learning is introduced to optimize the model's structured output capability and task performance. In the GRPO stage, the team jointly used several reward components, including format consistency reward, overall score reward, level accuracy reward, and QA accuracy reward. For checkpoints with better performance, the team continued training them with additional steps for further optimization.

For data processing, the team used the official PortraitCraft training data as the primary training source~\cite{sha2026portraitcraft}. Each training sample consists of an image input and a structured text output, covering the overall composition score, fine-grained composition attribute judgments, and image-related QA results. For image preprocessing, each input image is resized by scaling the shorter side to 640 while keeping the aspect ratio of the longer side unchanged. To increase the diversity of training samples, the team further constructed additional QA data based on the official training images. An external model was used to generate supplementary questions and answers offline, and this external model was only used for training data construction and did not participate in final inference or submission result generation.

During inference, the team adopted a test-time augmentation strategy. For the same input image, they performed multiple inference runs using different image scales, different prompt templates, and different option orders for the QA task. For the QA task, multiple prompts and different option orders were used to reduce the randomness introduced by a single prompting strategy, and the final answer was selected based on result consistency. For overall score and composition level prediction, outputs from different LoRA adapters and different input configurations were integrated to obtain a more stable final prediction. The final outputs were organized into the official submission format, including structured field formatting, label name normalization, QA option normalization, and final JSON file generation.

The team also summarized the main design ideas, functional roles, and experimental observations of different method variants, as shown in Tab.~\ref{tab:zte_chu_variants}.

\subsection{Track 2: Portrait Composition Generation}

\subsubsection{Team ContentU\&G}

Team ContentU\&G proposed a three-stage framework for Track 2 portrait composition generation. The method first contrasts each real training image with its teacher-generated counterpart to extract recurring failure modes, pairs each failure mode with a targeted edit instruction, and consolidates them into rule prompts that turn a VLM into a composition- and content-aware critic agent. The critic agent then iteratively corrects the closed-source teacher's pseudo images on the test prompts, producing supervision closer to the ground-truth distribution. Finally, the team assembles DPO preference pairs from both training and corrected test sides, uses human pose prompts as the ControlNet condition, and applies LoRA-based DPO training on the open-source Z-Image-Turbo backbone. The team reported a total score of 87.36.

As shown in Fig.~\ref{fig:contentug_pipeline}, the method consists of three stages: Prone-to-Error Extraction, Pseudo-Image Synthesis, and Pose-Conditioned Preference Distillation. In the first stage, Prone-to-Error Extraction contrasts each real training image with its teacher-generated counterpart to mine recurring failure modes and consolidates them, together with image-grounded edit instructions, into rule prompts that turn a VLM into a composition- and content-aware critic. In the second stage, Pseudo-Image Synthesis deploys this critic at test time to iteratively revise the closed-source teacher's outputs on the test prompts, producing supervision that lies closer to the real distribution. In the third stage, Pose-Conditioned Preference Distillation aggregates DPO preference pairs from both training and testing datasets to train a DPO-LoRA adapted to Z-Image-Turbo, incorporating a frozen pose-conditioned ControlNet branch for structural guidance.

\paragraph{Prone-to-Error Extraction.}
Let $D_{\mathrm{train}}=\{(x_i^*, c_i)\}_{i=1}^{N}$ denote the official training set, where $x_i^*$ is a real image and $c_i=(c_i^{\mathrm{comp}}, c_i^{\mathrm{cont}})$ is its pair of composition and content prompts. The team samples a subset of training examples and regenerates, for each sampled prompt, a counterpart with $G$ under the same condition:
\begin{equation}
\tilde{x}_i = G(c_i).
\end{equation}
Since $G$ is not free of compositional and content errors, the real image $x_i^*$ and its synthesized counterpart $\tilde{x}_i$ form a natural preference pair, with $x_i^*$ taken as the chosen sample and $\tilde{x}_i$ as the rejected one. These pairs provide a direct training signal for DPO, encouraging the student to follow the prescribed composition and content more faithfully than $G$ does.

The same pairs further act as a probe of where $G$ tends to fail. Conditioned on $c_i$, a vision-language model is prompted to list the visible discrepancies of $\tilde{x}_i$ relative to $x_i^*$, together with a corresponding edit instruction that describes how $\tilde{x}_i$ should be modified to recover the configuration of $x_i^*$. The descriptions are aggregated across the sampled subset into several failure modes and their representative edit instructions. The failure modes $F=\{f_1,\ldots,f_M\}$ encode generation pitfalls that $G$ tends to commit at the prompt level. Each $f_m$ is then realized as a single rule prompt $p_m$ that, together with its associated edit instruction template, drives the VLM to act as a composition- and content-aware critic. Given an image, its prompt $c$, and one $p_m$, the VLM returns a structured record:
\begin{equation}
\mathrm{VLM}(x,p_m,c)\rightarrow \{r_{\mathrm{det}}, r_{\mathrm{ins}}\},
\end{equation}
where $r_{\mathrm{det}}\in\{0,1\}$ indicates whether the corresponding pitfall is present in $x$, and $r_{\mathrm{ins}}$ is a concrete edit instruction grounded in $f_m$. The collection $P=\{p_1,\ldots,p_M\}$ turns the VLM into a critic agent that carries the prior distilled from $D_{\mathrm{train}}$ and is later reused to provide feedback on the teacher's pseudo images at test time.

\begin{figure*}[!p]
\centering
\includegraphics[width=\textwidth]{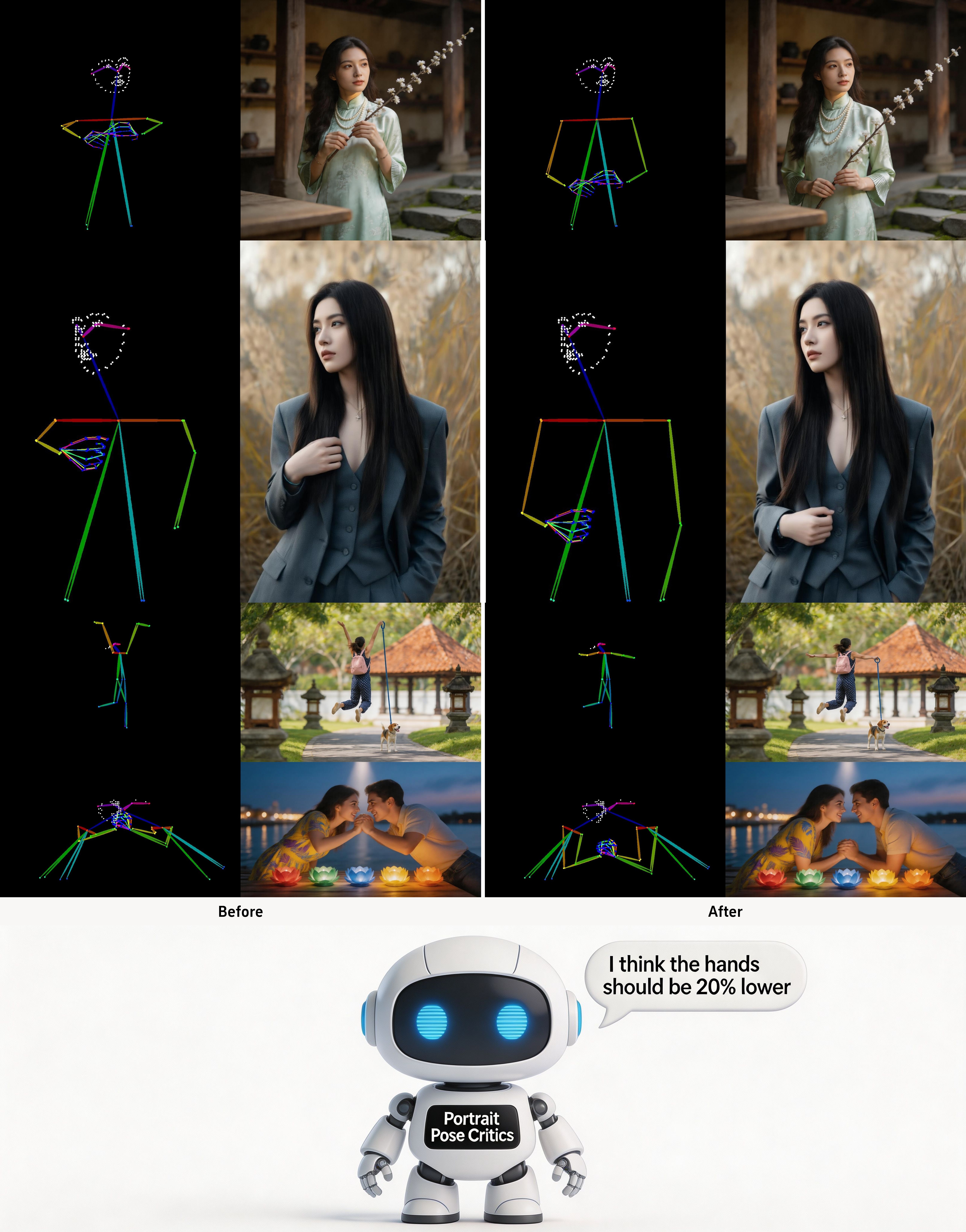}
\caption{Illustration of the pose-correction process guided by the Portrait Pose Critic.}
\label{fig:contentug_pose}
\end{figure*}

\begin{figure*}[!p]
\centering
\includegraphics[width=\textwidth]{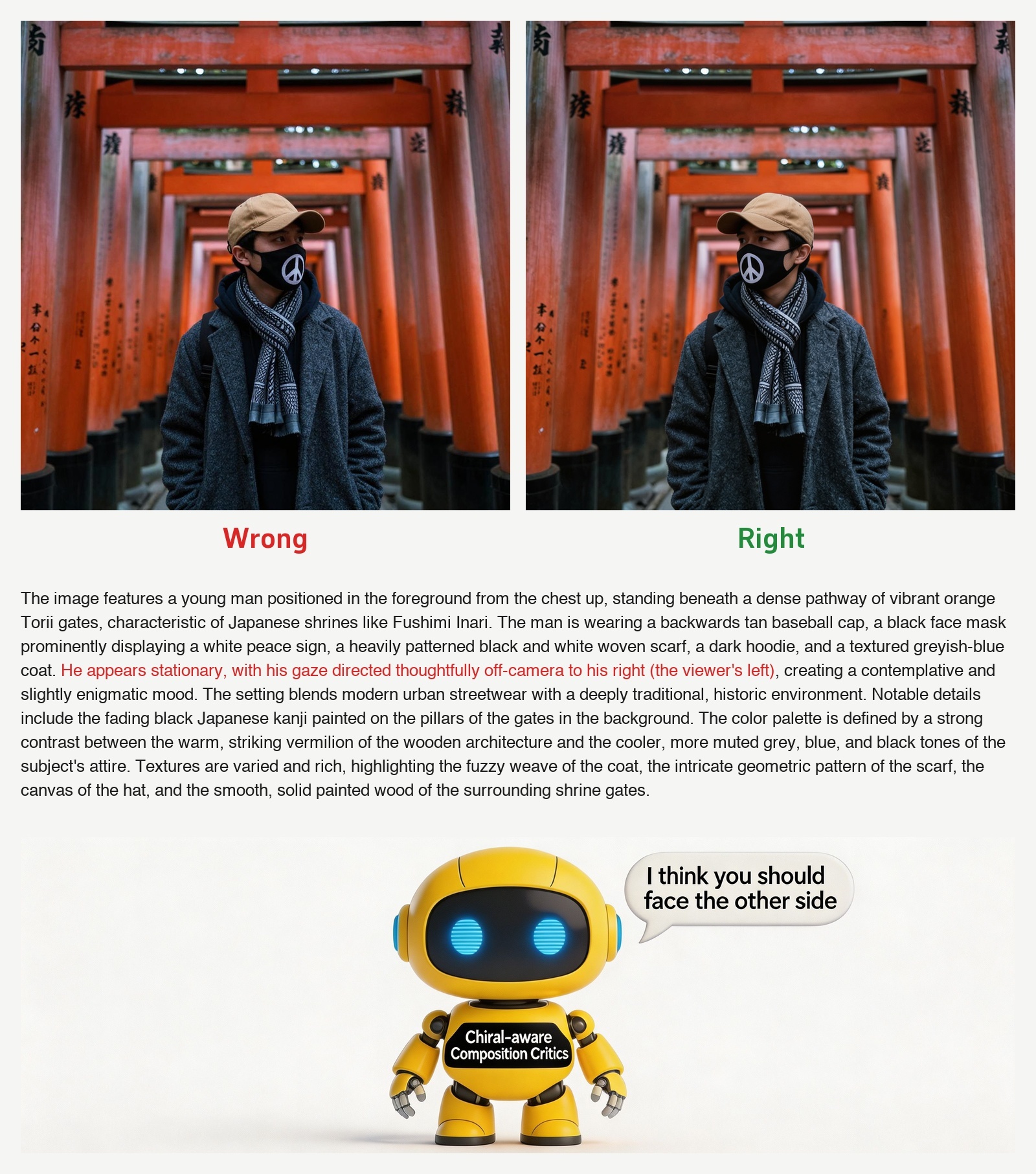}
\caption{Illustration of the orientation-correction process guided by the Chiral-aware Composition Critic.}
\label{fig:contentug_chiral}
\end{figure*}

\begin{figure*}[!p]
\centering
\includegraphics[width=0.7\textwidth]{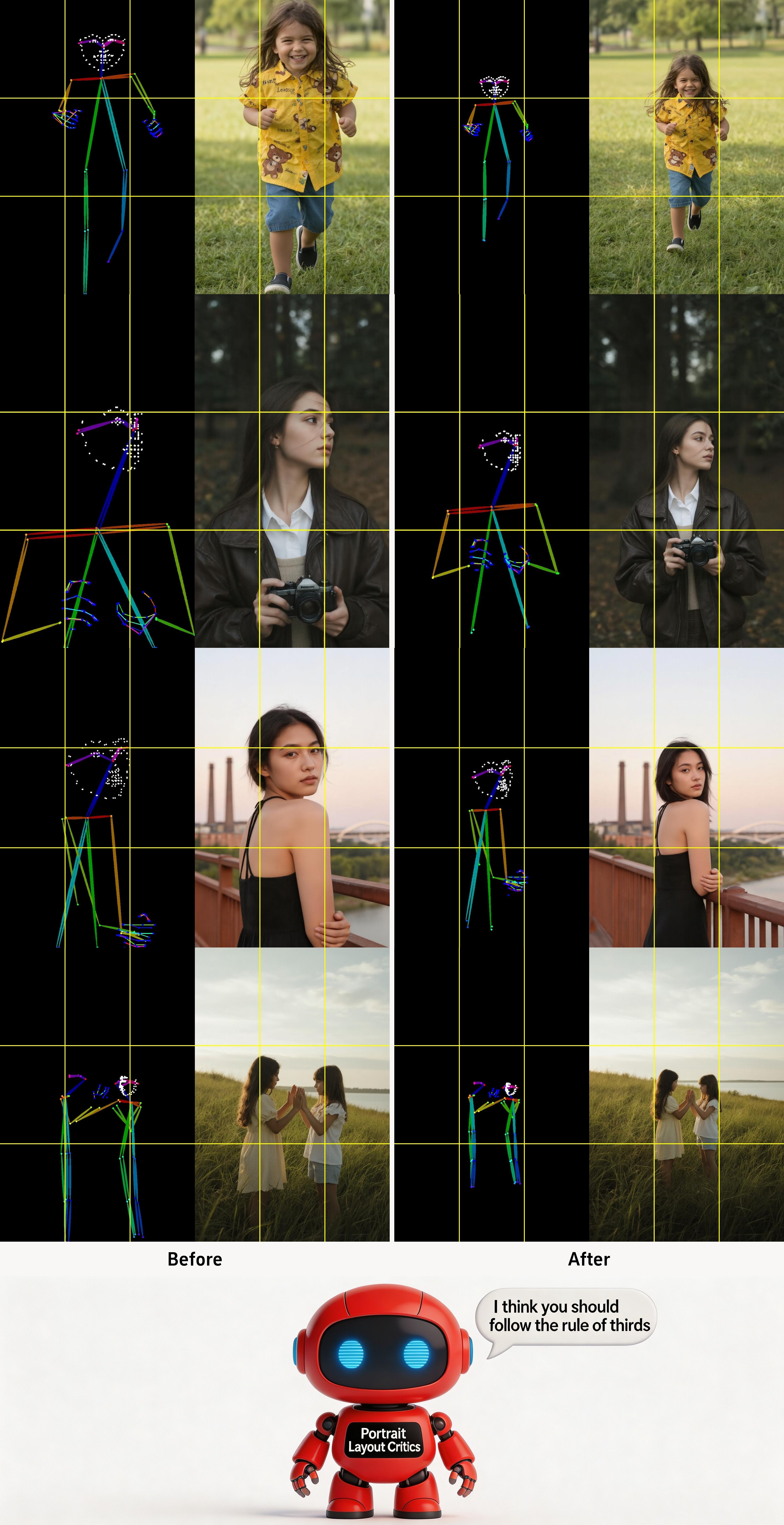}
\caption{Illustration of macro-compositional refinement guided by the Portrait Layout Critic.}
\label{fig:contentug_layout}
\end{figure*}

In practice, the team systematically identified and categorized recurring error-prone patterns inherent to state-of-the-art image generators. These failure modes include macro-level portrait compositional anomalies such as improper pose layout, as illustrated in Fig.~\ref{fig:contentug_pose}, fine-grained chiral orientation errors including left-right geometric confusion and gaze mismatches, as shown in Fig.~\ref{fig:contentug_chiral}, and rule-of-thirds violations, as shown in Fig.~\ref{fig:contentug_layout}. This structured taxonomy serves as the foundational prior for the downstream self-correction loop.

\paragraph{Pseudo-Image Synthesis.}
Given a test prompt $c_j$, the critic agent built in the previous stage is used to iteratively refine the pseudo image produced by $G$ so that it approaches the real distribution. Starting from the draft $\tilde{x}^{(0)}_j=G(c_j)$, at each iteration the critic inspects the current image with all rule prompts in $P$, identifies the remaining pitfalls, and returns a merged edit instruction $u^{(k)}$. Rather than introducing an additional editing model, the team lets $G$ revise its own previous output under this instruction:
\begin{equation}
\tilde{x}^{(k)}_j = G\left(\tilde{x}^{(k-1)}_j, u^{(k)}\right).
\end{equation}
The loop terminates as soon as the critic reports no remaining pitfall, and the converged image is denoted by $\hat{x}_j$. The draft $\tilde{x}^{(0)}_j$ and its corrected counterpart $\hat{x}_j$ form an additional preference pair on the test side, in which the critic-approved $\hat{x}_j$ is taken as the chosen sample and the original draft $\tilde{x}^{(0)}_j$ as the rejected one, providing test-distribution-aligned supervision for the DPO objective.

\paragraph{Pose-Conditioned Preference Distillation.}
The two preceding stages supply all of the training data. To couple every preference pair with an explicit compositional condition, the team attaches a 2D pose to each sample using an off-the-shelf pose estimator. For a training pair $(x_i^*,\tilde{x}_i)$, the pose $p_i^*$ is extracted from the real image $x_i^*$, while for a test-side pair $(\hat{x}_j,\tilde{x}^{(0)}_j)$, the pose $\hat{p}_j$ is extracted from the critic-corrected pseudo image $\hat{x}_j$. From the first stage, every training sample is cast as a DPO quadruple
\[
\left(c_i, x_i^{\mathrm{chosen}}=x_i^*, x_i^{\mathrm{rejected}}=\tilde{x}_i, p_i^*\right),
\]
in which the pose condition $p_i^*$ is shared by both sides so that the preference signal reflects image quality rather than differences in pose. From the second stage, every test prompt yields a quadruple
\[
\left(c_j, x_j^{\mathrm{chosen}}=\hat{x}_j, x_j^{\mathrm{rejected}}=\tilde{x}^{(0)}_j, \hat{p}_j\right),
\]
which provides prompt-conditioned supervision close to the test distribution after the bias of $G$ has been reduced by the prior $P$.

For the architecture, the team adopts a ``Base Model + DPO-LoRA + ControlNet'' design. The LoRA module, inspired by methods like~\cite{f162025zimageflowdpo}, is attached to the base model and is designed to absorb the preference signal mined from the data pipeline. The ControlNet branch turns the pose keypoints $p_i^*$ and $\hat{p}_j$ into explicit spatial guidance for composition. This converts implicit textual cues about layout into a structured input and improves compositional stability in complex scenes.

The LoRA is trained with DPO on the quadruples produced by the previous stages, with the corresponding pose condition injected through the ControlNet branch. Let $\pi_{\theta}$ denote the LoRA-augmented policy and $\pi_{\mathrm{ref}}$ its frozen reference. Following~\cite{rafailov2024dpo}, the DPO objective is:

\begin{equation}
\begin{aligned}
\mathcal{L}_{\mathrm{DPO}}
= -\log \sigma \Bigg(
& \beta \log
\frac{\pi_{\theta}(x_i^* \mid c_i,p_i^*)}
{\pi_{\mathrm{ref}}(x_i^* \mid c_i,p_i^*)} \\
& -
\beta \log
\frac{\pi_{\theta}(\tilde{x}_i \mid c_i,p_i^*)}
{\pi_{\mathrm{ref}}(\tilde{x}_i \mid c_i,p_i^*)}
\Bigg).
\end{aligned}
\end{equation}

\clearpage
\clearpage
\clearpage

where $\sigma(\cdot)$ is the sigmoid function and $\beta$ controls the deviation from the reference. Since $p_i^*$ appears on both sides of the contrast, the loss isolates image-quality preference from pose conditioning and prevents the model from explaining the preference signal away through pose mismatch.

\paragraph{Implementation Details.}
The team adopts Z-Image-Turbo~\cite{zimage2025} as the base model, which offers a favorable trade-off between overall performance and parameter scale. For the control branch, they use Z-Image-Turbo-Fun-Controlnet-Union-2.1~\cite{pai2025controlnet}. Although this union model supports multiple control modalities, including Canny, Depth, Pose, MLSD, HED, Scribble, and Grayscale, only the Pose channel is activated in their scheme, and the ControlNet weights are kept frozen during training. The DPO-LoRA is trained for 19,200 steps on $8\times$A100 GPUs. The team observed that DPO training of the LoRA partially degrades the inherent distillation property of Z-Image-Turbo; therefore, they set the number of inference steps to 40. They also experimented with the distill patch of~\cite{diffsynth2025distillpatch} to restore the distillation behavior, but it led to an overall performance drop and was not used in the final system.

\subsubsection{Team elephant}

\begin{figure*}[!t]
\centering
\includegraphics[width=\textwidth]{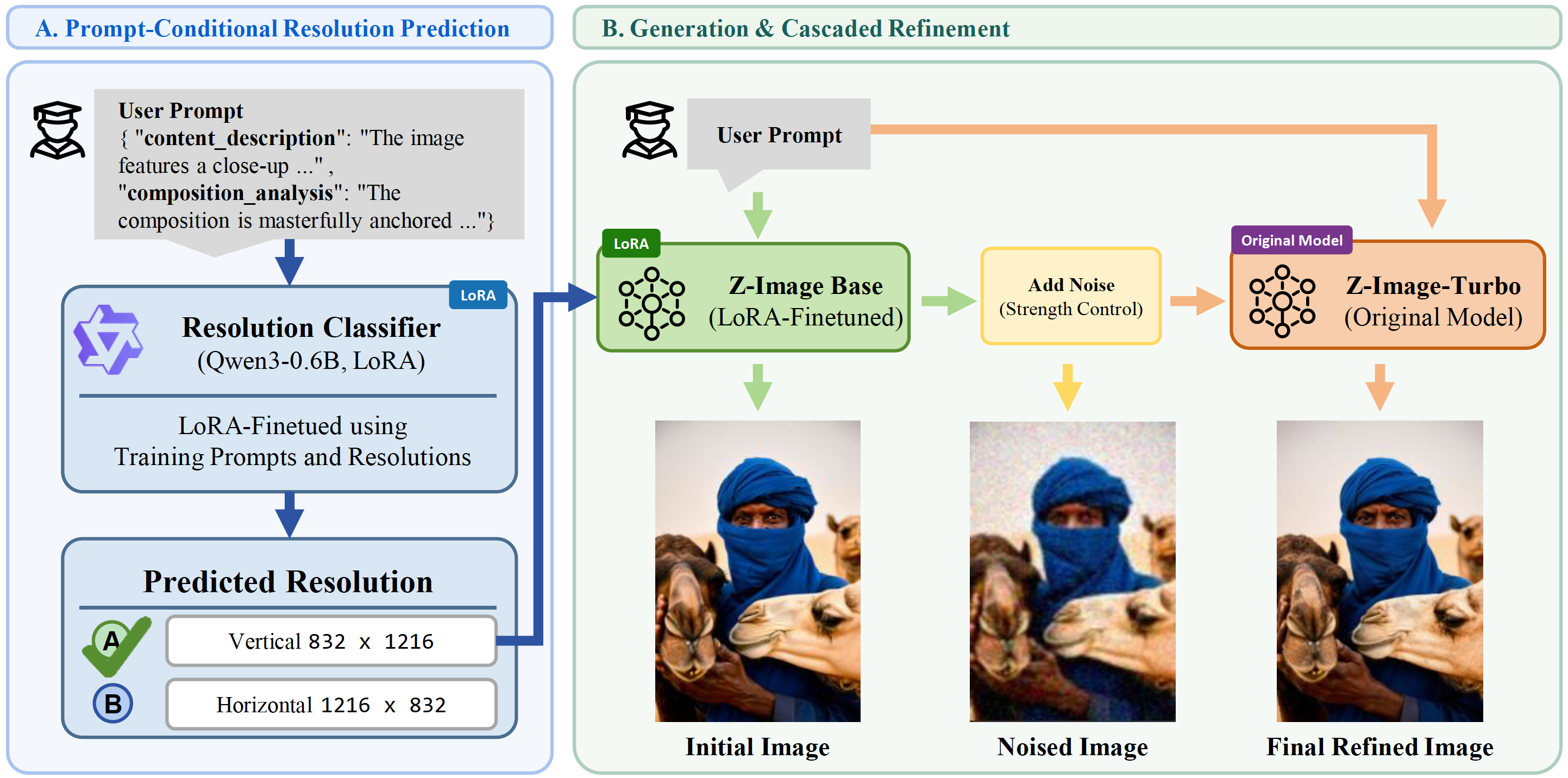}
\caption{Overview of the three-stage inference pipeline of Team elephant.}
\label{fig:elephant_pipeline}
\end{figure*}

Team elephant presented a solution for PortraitCraft Challenge Track 2, which formulates portrait composition generation as a conditional image generation problem at the 3rd AI for Visual Arts Workshop at CVPR 2026~\cite{bhattacharjee2026ai4va}. Their approach is built upon Z-Image~\cite{zimage2025}, a 6B-parameter image generation foundation model based on the Scalable Single-Stream Diffusion Transformer (S3-DiT) architecture. The team adapts Z-Image to the portrait composition domain through Low-Rank Adaptation (LoRA)~\cite{hu2022lora}, combining the official challenge training data with a curated supplementary dataset sourced from Unsplash~\cite{unsplash2023datasets}. To further improve composition quality, they train a lightweight Qwen3-0.6B~\cite{qwen2025qwen3} classifier that predicts the optimal aspect ratio conditioned on the input textual description. At inference time, the method employs a three-stage pipeline: aspect ratio prediction, composition-conditioned generation via the LoRA-adapted Z-Image, and cascaded refinement through Z-Image-Turbo~\cite{zimage2025}.

\paragraph{Training.}
For the image generation backbone, the team adopts Z-Image~\cite{zimage2025}. Z-Image is a 6B-parameter generative model built upon the S3-DiT architecture, which concatenates text tokens, visual semantic tokens, and image VAE tokens at the sequence level as a unified input stream. This design enables dense cross-modal interaction at every transformer layer. The model exhibits strong photorealistic generation capability, aesthetic versatility, and prompt adherence, making it a suitable foundation for composition-aware portrait generation.

For fine-tuning, the team uses LoRA~\cite{hu2022lora} rather than full fine-tuning, since full fine-tuning is computationally prohibitive for a 6B-parameter model and risks catastrophic forgetting of the pretrained generative prior. LoRA freezes the pretrained weights and injects trainable low-rank decomposition matrices into the attention layers of the transformer. The team sets the LoRA rank to $r=32$ to balance model capacity for learning composition-specific patterns and regularization against overfitting on limited training data. The LoRA parameters are optimized using AdamW~\cite{loshchilov2019adamw} with a learning rate of $1\times10^{-4}$, a batch size of 8, and 8,000 training iterations.

The training dataset comprises two complementary sources. The first source consists of 4,500 portrait images paired with structured textual descriptions provided by the challenge organizers, covering composition attributes such as subject prominence, spatial organization, pose quality, visual center, negative space, and overall composition style. The second source is a supplementary dataset curated from the publicly available Unsplash dataset~\cite{unsplash2023datasets}. The team first filters the Unsplash dataset to retain only images containing human subjects using person detection. Then, each candidate image is scored using the Improved Aesthetic Predictor~\cite{schuhmann2022aesthetic}, a CLIP-based model trained on the AVA dataset~\cite{murray2012ava}, and images with an aesthetic score exceeding 5.5 are retained. The team randomly samples 10,000 images from the filtered pool and generates structured composition-oriented textual descriptions for each image using Qwen3.6~\cite{qwen2026qwen36}, a 27B-parameter multimodal model supporting vision-language understanding. In total, the training set comprises approximately 14,500 image-text pairs.

\paragraph{Aspect Ratio Prediction.}
The team observes that the aspect ratio of the generated image significantly impacts perceived composition quality. For example, full-body portraits with vertical leading lines favor portrait orientation, while wide multi-subject group scenes are better suited to landscape orientation. To address this, they train a dedicated aspect ratio classifier based on Qwen3-0.6B~\cite{qwen2025qwen3}. Given a structured composition description as input, the model predicts a binary orientation label, \texttt{vertical} or \texttt{horizontal}, which is mapped to fixed resolutions of $832\times1216$ and $1216\times832$, respectively. The classifier is fine-tuned with orientation labels derived from the aspect ratios of the training images.

\paragraph{Inference.}
Given a test composition description, the inference pipeline proceeds in three sequential stages, as shown in Fig.~\ref{fig:elephant_pipeline}. In Stage 1, the input description is passed to the fine-tuned Qwen3-0.6B aspect ratio classifier, which predicts whether the target portrait should adopt the vertical resolution $832\times1216$ or the horizontal resolution $1216\times832$. In Stage 2, the structured composition description, together with the predicted resolution, is fed into the LoRA-adapted Z-Image model. The model performs text-conditioned denoising over 50 inference steps with a classifier-free guidance scale of 4.0 to synthesize a portrait image that realizes the specified composition attributes, including subject placement, spatial organization, visual flow, and aesthetic style.

In Stage 3, the team applies Z-Image-Turbo~\cite{zimage2025} as a cascaded refinement stage to enhance photorealistic quality and fine-grained detail. Z-Image-Turbo is a few-step distilled variant of Z-Image obtained via Decoupled Distribution Matching Distillation combined with reinforcement learning-based reward post-training, enabling high-fidelity synthesis with only 8 function evaluations. The team employs Z-Image-Turbo in an image-to-image fashion following the SDEdit paradigm~\cite{meng2022sdedit}, with a denoising strength of 0.3. This conservative strength preserves the compositional structure from Stage 2 while improving surface-level realism, including skin texture, lighting coherence, and detail sharpness.

\paragraph{Experiments.}
The team conducts an ablation study by progressively incorporating modules into the pipeline. The evaluation is performed on the official PortraitCraft Challenge Track 2 test set, and three metrics are reported following the challenge protocol: Content Score, Composition Score, and Total Score. The results are summarized in Tab.~\ref{tab:elephant_ablation}. LoRA fine-tuning provides the largest improvement, increasing the Total Score from 70.9 to 83.5. Cascaded refinement provides a marginal improvement from 83.5 to 83.6. Adding aspect ratio prediction further improves the Composition Score from 82.6 to 84.0 and increases the Total Score to 84.6, while the Content Score remains 86.0.

\begin{table}[t]
\centering
\caption{Ablation study on the PortraitCraft Track 2 test set reported by Team elephant.}
\label{tab:elephant_ablation}
\begin{adjustbox}{width=\linewidth}
\begin{tabular}{lccc}
\toprule
Configuration & Content$\uparrow$ & Composition$\uparrow$ & Total$\uparrow$ \\
\midrule
(a) Z-Image (no fine-tuning) & 79.5 & 67.3 & 70.9 \\
(b) + LoRA SFT & 85.9 & 82.5 & 83.5 \\
(c) + Cascaded refinement & 86.0 & 82.6 & 83.6 \\
(d) + Aspect ratio prediction (full) & \textbf{86.0} & \textbf{84.0} & \textbf{84.6} \\
\bottomrule
\end{tabular}
\end{adjustbox}
\end{table}

\subsubsection{Team sky}

\begin{figure*}[!t]
\centering
\begin{minipage}{0.24\textwidth}
\centering
\includegraphics[width=\linewidth]{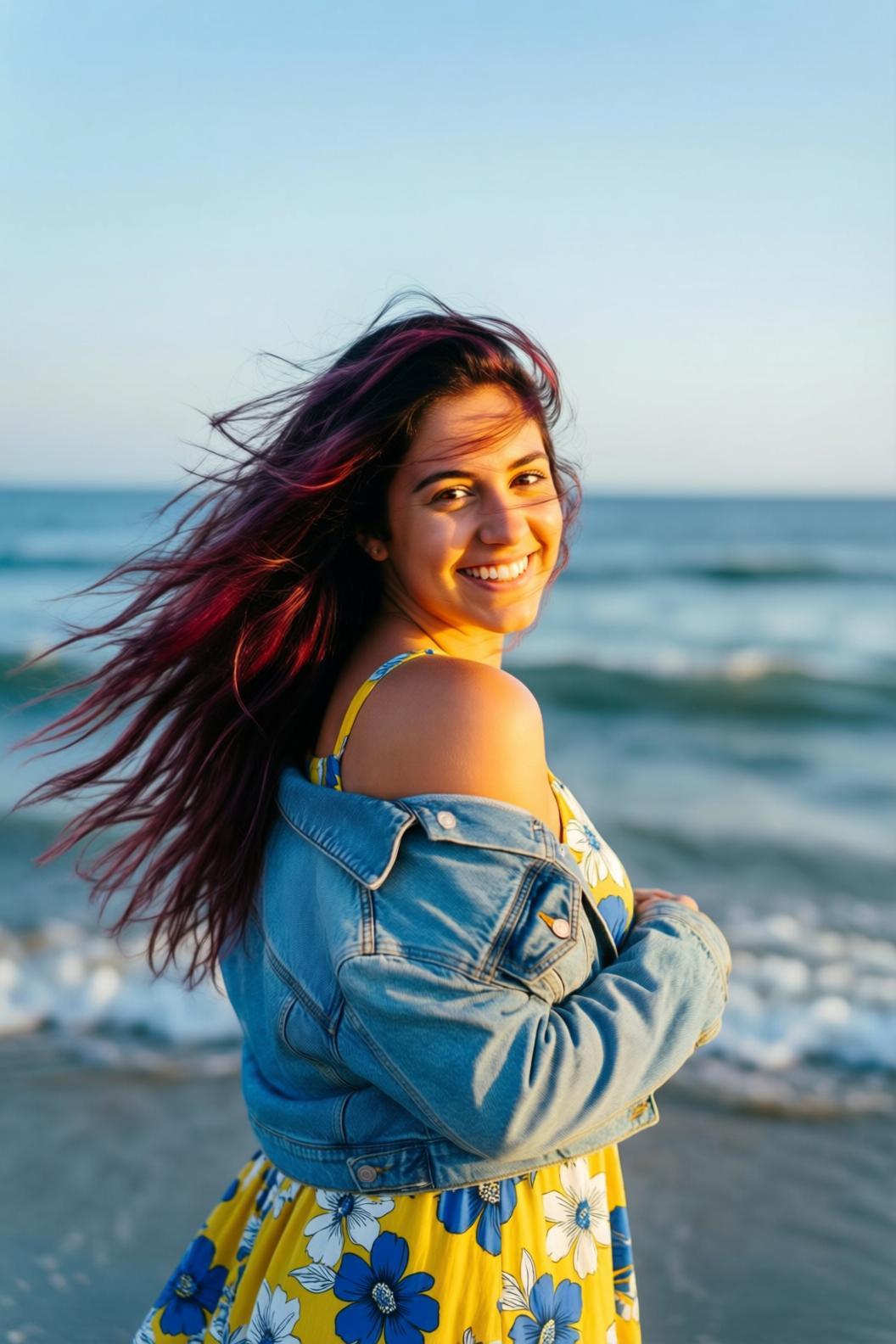}
\caption*{Square portrait canvas}
\end{minipage}
\hfill
\begin{minipage}{0.24\textwidth}
\centering
\includegraphics[width=\linewidth]{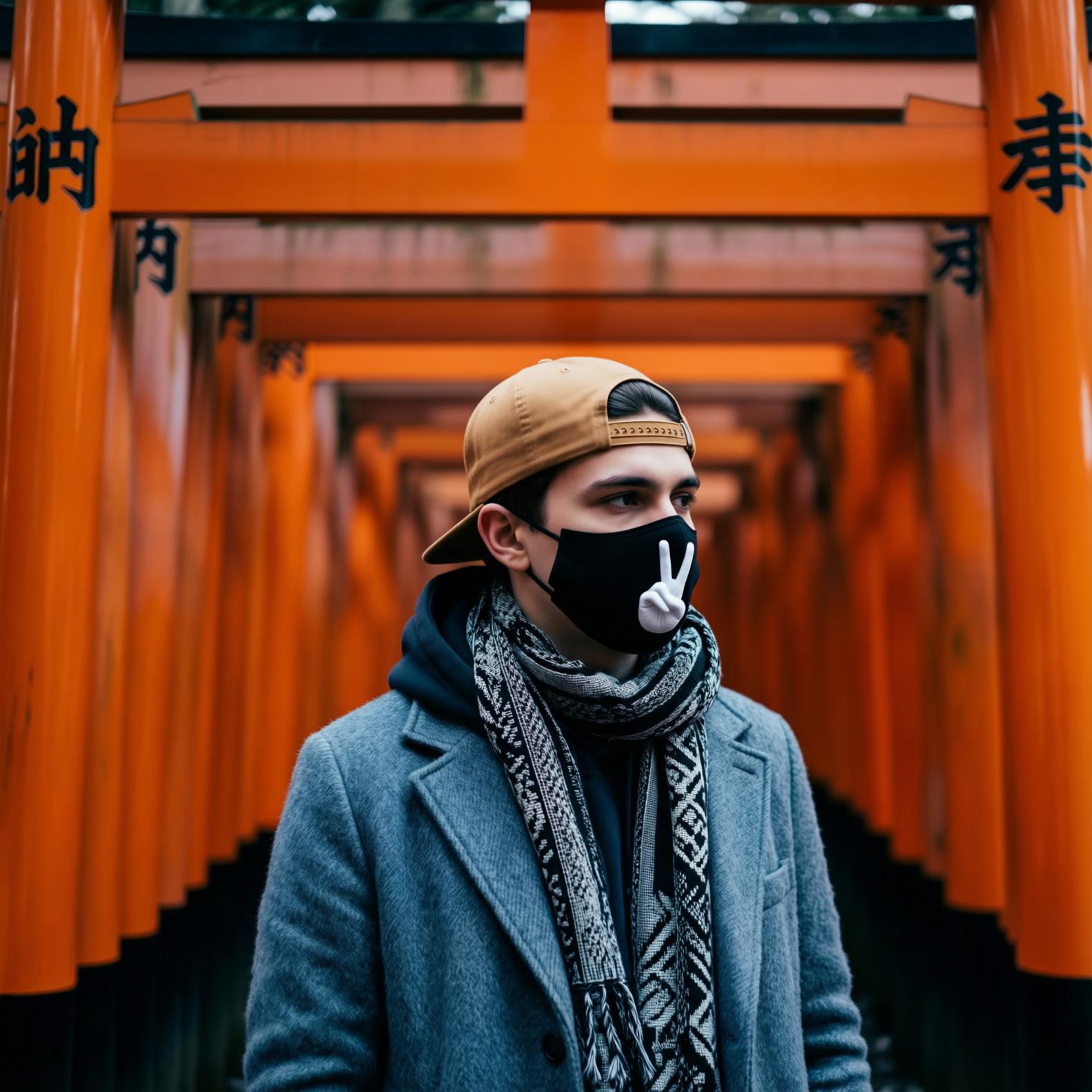}
\caption*{Vertical portrait canvas}
\end{minipage}
\hfill
\begin{minipage}{0.24\textwidth}
\centering
\includegraphics[width=\linewidth]{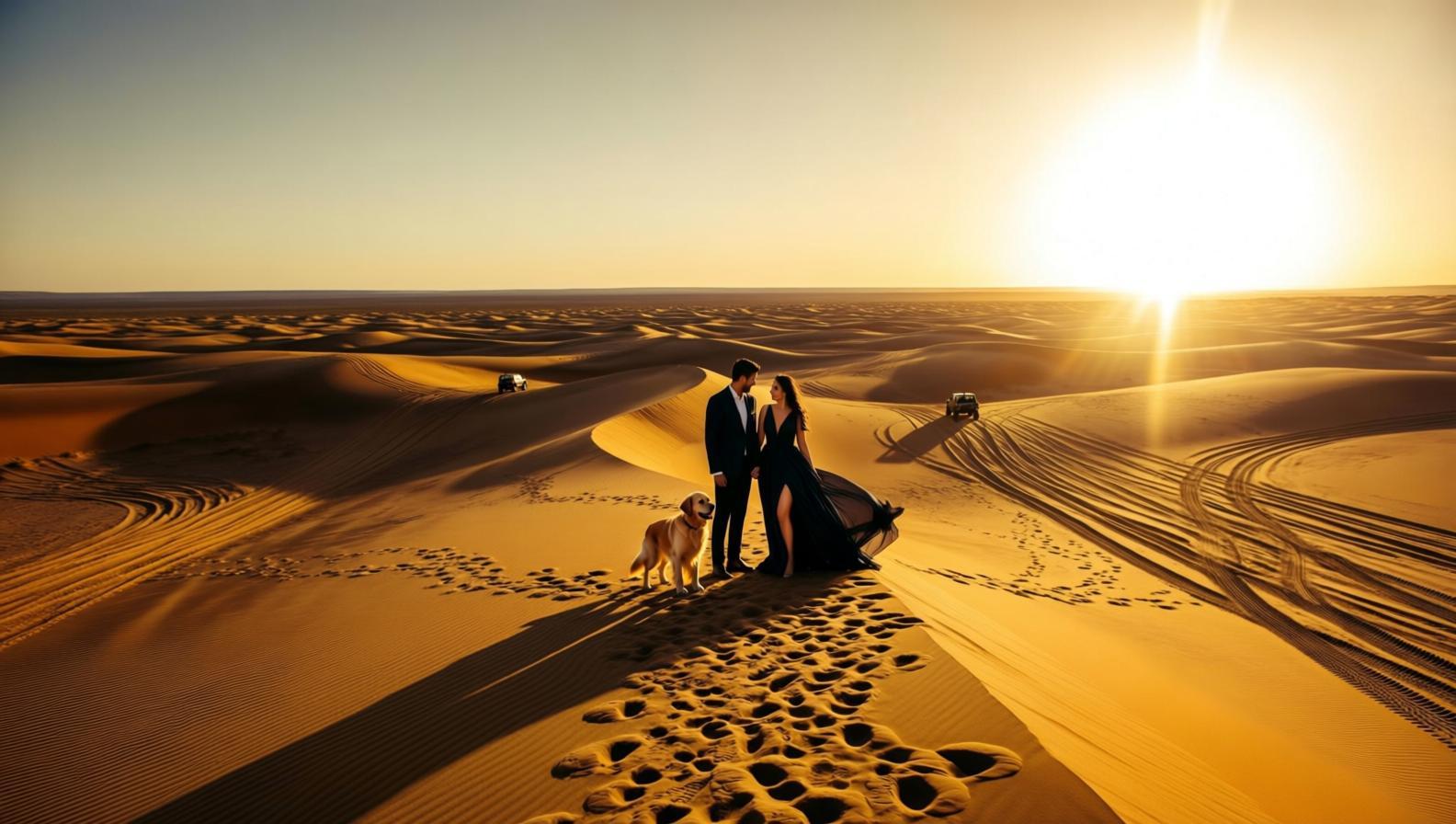}
\caption*{Horizontal environmental portrait}
\end{minipage}
\hfill
\begin{minipage}{0.24\textwidth}
\centering
\includegraphics[width=\linewidth]{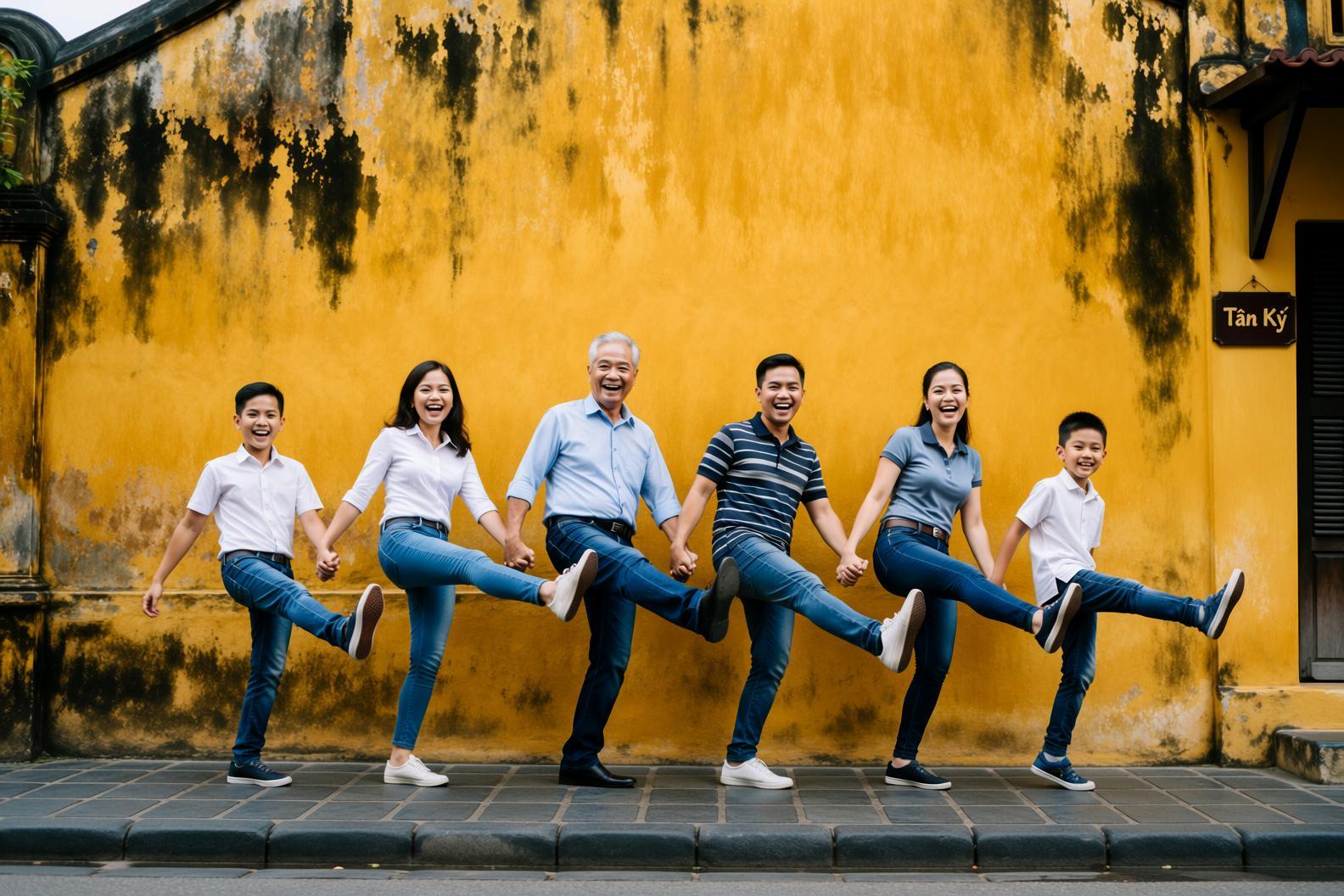}
\caption*{Wide environmental composition}
\end{minipage}
\caption{Qualitative examples from Team sky showing different canvas choices selected by the adaptive canvas policy.}
\label{fig:sky_examples}
\end{figure*}

Team sky proposed a solution for PortraitCraft Track 2 based on Qwen-Image-2512. The core idea is to combine a portrait-composition fine-tuned Qwen-Image-2512 model with a prompt-conditioned adaptive canvas policy. The generation model improves visual quality and prompt-to-layout alignment, while the canvas policy selects a suitable aspect ratio before generation.

The solution focuses on two complementary components: a portrait-composition fine-tuned generation model and a prompt-conditioned adaptive canvas policy. The generation model synthesizes visually coherent portrait images, while the canvas policy selects the generation aspect ratio before sampling. The team does not use a fixed $1{:}1$ canvas for all images, since close-up and centered portraits often work well on square canvases, full-body portraits benefit from vertical layouts, and environmental portraits or scenes with roads, coastlines, leading lines, or large negative space often need horizontal or wide canvases.

\paragraph{Training Data and Fine-Tuning.}
The team fine-tuned Qwen-Image-2512 using the official 4,500 PortraitCraft training samples together with an additional private portrait aesthetic-composition dataset curated by the team. The private data focuses on portrait layout, aesthetic framing, human-subject placement, environmental context, lighting balance, and composition consistency. The team compared LoRA fine-tuning and full-parameter fine-tuning under the same inference settings. Full-parameter fine-tuning was selected for the final submission because it performed better for this task, especially in aesthetic quality, composition stability, and prompt-to-layout alignment.

\paragraph{Adaptive Canvas Policy.}
Instead of using a fixed $1{:}1$ canvas for all images, the team uses a prompt-conditioned adaptive canvas policy. The policy reads the input prompt and a learned policy state, then outputs a canvas size before image generation. The longer side is normalized to 1584 pixels, while the shorter side is selected from a compact set of portrait-friendly aspect ratios. The policy was optimized on the training set through an iterative evolutionary-search procedure, which adjusted keyword weights, decision thresholds, and candidate aspect-ratio choices. This allows the inference system to preserve the intended spatial structure for different prompt types, including square portraits, full-body vertical portraits, and horizontal environmental portraits. If the prompt evidence is not confident enough to select a non-square layout, the policy conservatively falls back to a square $1584\times1584$ canvas.

For reproducibility, the team released the final learned policy state together with the inference code. Reviewers can recover the same canvas selection used by the submission. For unseen prompts, the implementation falls back to a deterministic prompt-only rule policy.

\paragraph{Inference Pipeline.}
The final inference pipeline first applies the adaptive canvas selector to each prompt, then generates the image using the released PortraitCraft Track 2 checkpoint. The output images are saved with their original task filenames and packaged as a flat zip file. The input consists of the text description and composition analysis. The adaptive canvas policy selects the width and height with the longest side set to 1584. The Qwen-Image-2512 model with the PortraitCraft checkpoint then generates the image using 50 sampling steps, CFG scale 4.0, and a fixed seed.

\begin{table}[!t]
\centering
\caption{Inference configuration of Team sky.}
\label{tab:sky_inference_config}
\begin{adjustbox}{width=\linewidth}
\begin{tabular}{lc}
\toprule
Item & Setting \\
\midrule
Base model & Qwen-Image-2512 \\
Checkpoint & portraitcraft-track2.safetensors \\
Sampling steps & 50 \\
CFG scale & 4.0 \\
Seed & 346346 \\
Adaptive canvas longest side & 1584 pixels \\
\bottomrule
\end{tabular}
\end{adjustbox}
\end{table}

\begin{figure*}[!t]
\centering
\includegraphics[width=\textwidth]{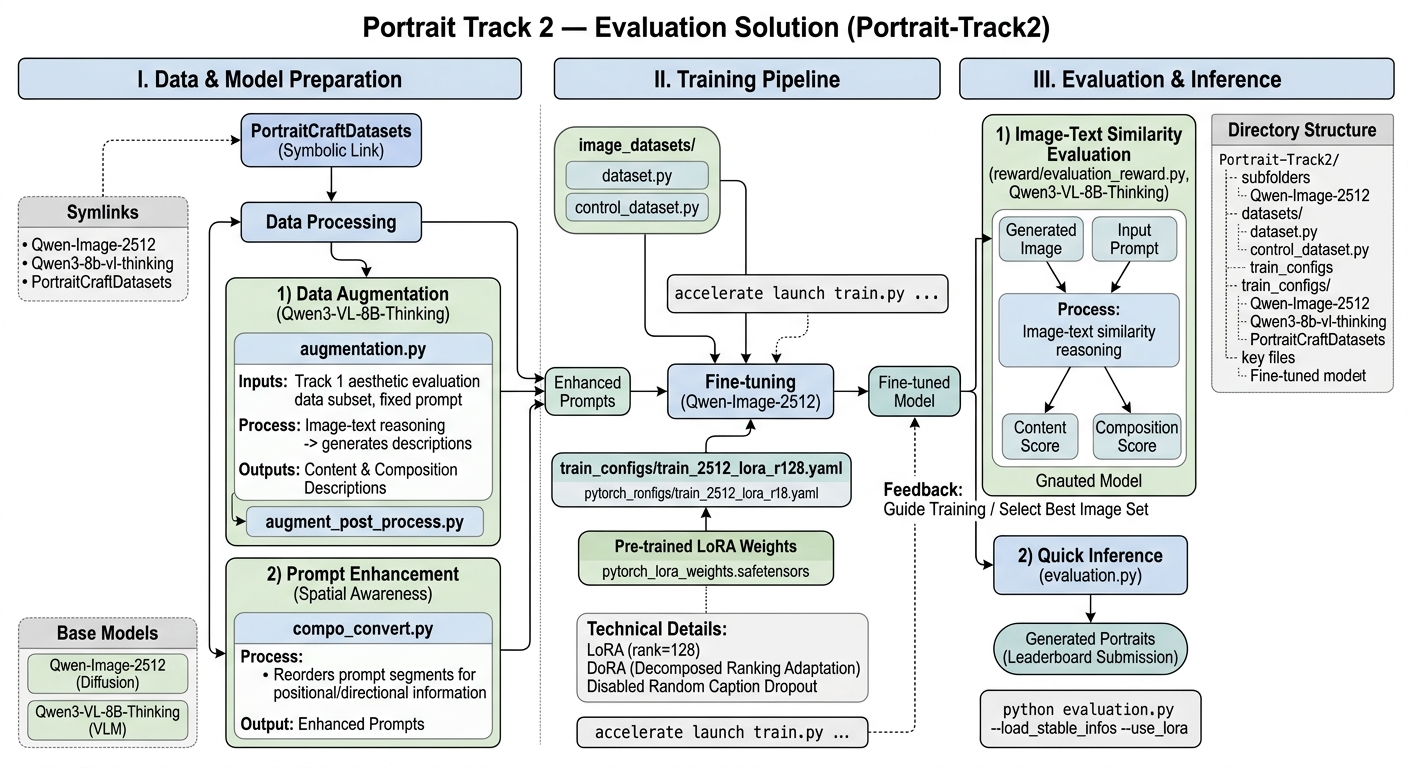}
\caption{Overview of the Track 2 evaluation solution proposed by Team PCU-vRobotit@BUPT.}
\label{fig:pcu_bupt_overview}
\end{figure*}

\paragraph{Qualitative Examples.}
The team provided qualitative examples to illustrate how the adaptive canvas policy supports different composition needs. Square layouts are used for centered portraits, vertical layouts preserve human-body framing and breathing room, horizontal layouts support environmental context and directional visual flow, and wider layouts support roads, coastlines, leading lines, and negative space, as shown in Fig.~\ref{fig:sky_examples}.

\paragraph{Summary.}
The final solution uses a full-parameter fine-tuned Qwen-Image-2512 checkpoint and a prompt-conditioned adaptive canvas policy. The training data combines the official PortraitCraft training set with private portrait aesthetic-composition data. The adaptive canvas policy was optimized on the training data and released with the code to make canvas selection reproducible. This design matches the canvas to the prompt's spatial intent before image synthesis, rather than forcing every image into the same square frame.

\subsubsection{Team PCU-vRobotit@BUPT}

Team PCU-vRobotit@BUPT proposed a Track 2 portrait image synthesis solution based on Qwen-Image-2512. Their solution fine-tunes Qwen-Image-2512 using LoRA and leverages Qwen3-VL-8B-Thinking for both data augmentation and image-text similarity evaluation. The overall pipeline is shown in Fig.~\ref{fig:pcu_bupt_overview}.

\paragraph{Data Augmentation and Prompt Enhancement.}
The team augments a subset of the Track 1 image aesthetic evaluation data. Using the image-text reasoning capability of Qwen3-VL-8B-Thinking, they generate content and composition descriptions for each image given a fixed prompt. They also observe that text-to-image models are insensitive to positional and directional information in default prompts. To address this, they reorder certain prompt segments so that the model pays more attention to spatial relationships, improving generation quality.

\paragraph{Training.}
The team fine-tunes the base model Qwen-Image-2512 using LoRA. Due to the large model size, they disable the random caption dropout mechanism, use LoRA rank 128, and adopt DoRA (Decomposed Ranking Adaptation) for better performance. The team reports that experimental results confirm this configuration yields the best results among all tested setups.

\paragraph{Image-Text Similarity Evaluation.}
The team uses Qwen3-VL-8B-Thinking to build an image evaluation system. For each generated image, the model performs image-text similarity reasoning against its input prompt and outputs a content score and a composition score. These scores guide the training direction and help select the best image set, bypassing the limited number of website submissions.

\paragraph{Inference.}
For inference, the team uses the fine-tuned LoRA model with the provided evaluation script. The leaderboard submission can be reproduced by loading stable information and enabling LoRA during inference. The final generated portraits are used for leaderboard submission.

\section{Conclusion}
This report presented the 1st PortraitCraft Challenge, held as an official CVPR 2026 workshop competition. The challenge provided a comprehensive benchmark for portrait composition understanding and generation, attracting a total of 295 teams from universities, research institutions, and industry worldwide. The submitted solutions demonstrated strong performance on both tracks, significantly advancing the state of the art in structured portrait composition analysis and controllable portrait generation under explicit composition constraints. Strategies targeting fine-grained attribute reasoning, visual question answering, and composition-aware generation have shown great potential for future research. The PortraitCraft benchmark and challenge results provide a standardized, reproducible platform that will foster further progress in portrait aesthetics, interpretable composition assessment, and composition-guided image synthesis.
% 强制输出所有浮动体（图、表等）

{
    \small
    \bibliographystyle{ieeenat_fullname}
    \bibliography{main}
}

% WARNING: do not forget to delete the supplementary pages from your submission 
% \input{sec/X_suppl}

\end{document}